\documentclass[journal]{journal}
\ifCLASSINFOpdf
\else
\fi

\usepackage{graphicx}
\graphicspath{ {./images/} }

\begin{document}
\title{Deep Learning based, end-to-end metaphor detection  in Greek language  with Recurrent and Convolutional Neural Networks}
%
%
%  names and IEEE memberships

\author{Konstantinos~Perifanos \IEEEauthorrefmark{1}
        and~Eirini~Florou \IEEEauthorrefmark{2}
        and~Dionysis~Goutsos \IEEEauthorrefmark{3}\\% <-this % stops a space\\
Department of Linguistics, National and Kapodistrian University of Athens, Greece\\
\IEEEauthorrefmark{1}kperifanos@phil.uoa.gr,
\IEEEauthorrefmark{2}eirini.florou@gmail.com,
\IEEEauthorrefmark{3}dgoutsos@phil.uoa.gr,

}


\markboth{Journal of \LaTeX\ Class Files,~Vol.~6, No.~1, January~2007}%
{Shell \MakeLowercase{\textit{et al.}}: Bare Demo of IEEEtran.cls for Journals}



\maketitle
\thispagestyle{empty}

\begin{abstract}
%\boldmath
This paper presents and benchmarks a number of end-to-end Deep Learning based models for metaphor detection in Greek. We combine Convolutional Neural Networks and Recurrent Neural Networks with representation learning to bear on the metaphor detection problem for the Greek language. The models presented achieve exceptional accuracy scores, significantly
improving the previous state of the art results, which had already achieved accuracy 0.82. Furthermore, no special preprocessing, feature engineering or linguistic knowledge is used in this work. The methods presented achieve accuracy of 0.92 and F-score 0.92 with Convolutional Neural Networks (CNNs) and bidirectional Long Short Term Memory networks (LSTMs). Comparable results of  0.91 accuracy and  0.91 F-score are also achieved with bidirectional Gated Recurrent Units (GRUs) and Convolutional Recurrent Neural Nets (CRNNs).  The models are trained and evaluated only on the basis of the training tuples, the sentences and their labels. The outcome is a state of the art collection of metaphor detection models, trained on limited labelled resources, which can be extended to other languages and similar tasks.
\end{abstract}


% \fi
\IEEEpeerreviewmaketitle

\section{Introduction}

Metaphor as a figure of speech has a widespread presence in any form of communication either oral or written. According to Steen \cite{steen2010metaphor} data analysis shows that, on average, one in every seven and a half lexical units in the corpus is related to metaphor

However, it is difficult to clearly define the boundaries that separate metaphor from literal uses, as well as metaphor from other figures of speech.  The difficulty of clearly establishing a theoretical background for metaphor justifies the variety of NLP systems that aim at automatically between distinguishing between metaphorical and literal meanings of a word or phrase. This difficulty is further exacerbated if we take into account the limitations of Greek as regards resources and tools for metaphor detection; thus, we can conclude that the development of neural language models is necessary for the automatic differentiation between literal and metaphorical meaning of phrases that are part of an authentic and non-annotated Greek corpus. For these reasons, our attempt here is based on the principles of distributional semantics so as to determine the relations of a word with its linguistic context and to group semantic similarities of linguistic items based on distributional properties rather than any connections of the certain term and its related concepts. Distributional semantics  have been paramount in shifting research interest
towards neural language models, which can attribute hidden statistical characteristics of the distributed representations of word sequences in natural language. Therefore, a serious problem such as the automatic detection of metaphors and their differentiation from literal uses can be dealt with the development of neural language models.

\section{Previous work}
The computational identification and interpretation of metaphors have been based on a variety of computational tools like  statistical models \cite{met_dect_term_relev}, word taxonomies \cite{met_probabilities}, clustering \cite{Birke}, logistic regresion \cite{metaphor_id,dunn2014measuring} or  generative statistal models such as Latent Dirichlet Allocation (LDA) \cite{blei2003latent}.  As has happened with many linguistic phenomena, computational approaches to metaphor are now based on neural models and take advantage of the benefits of representation learning \cite{bengio2013representation}, and more specifically distributed representations, also known as word embeddings  \cite{koper-etal-2017-ims,rei2017grasping}. The neural models for metaphor detection include Long Short Term Memory (LSTMs) and Conditional Random Fields (CRFs) \cite{lafferty2001conditional}, which perform better with the contribution of linguistic features like  the Wordnet, POS tags or clustering.

The omnipresence of metaphor in all types of Greek texts had initially guided our research interest to an alternative approach to automatic metaphor detection, following the principles of distributional semantics and without the requirement of access to linguistic resources and tools  or exprensive and time-consuming manual annotation. This approach was based on neural language models and had taken into account the context of each term in order to identify its function and uses without explicitly employing any connections between this word and its related concepts. Neural language models offer the opportunity to a language which is poor of linguistic resources and tools to overpass the problem of calculating the semantic relevance between phrases. Taking advantage of the benefits of distributional semantics we substituted the semantic comparison of terms with a numerical comparison of their distributional representation in vector space. Through this comparison we were able to identify the literal or metaphorical function of words in a specific context. This first approach of metaphor detection in Greek texts is our baseline. However, we strived for improving the procedure  of metaphor detecion and for this reason we took into account state-of-the-art Deep Learning based models such as Convolutional and Recurrent Neural Networks in order to achieve the prediction of the metaphoricity of every word in a running text.

\section{Deep Learning for text classification}

Recent advances in Neural Networks and Transfer learning have been sussefully applied to Natural Language Processing. More specifically, Convolutional Neural Networks (CNNs, ConvNets) as well as Recurrent Neural Network (RNNs) architectures, have been applied to text classificiation problems, such as Named Entity Recognition, Part-of-Speech tagging, Semantic Role Labeling etc. \cite{kim2014convolutional,tang2015document,wang2016combination,collobert2011natural}

Training models with RNNs and CNNs from scratch typically requires a vast amount of labelled data, which is generally a time consuming and expensive process.

We tackle this by using transfer learning, and more specifically by using pre-trained word embeddings and allowing the model to fine-tune the first layer of the network (the embedding layer) as part of the training process. The term \textit{embeddings} refers to compact, continuous representations of words in a $D$-dimensional space and has emerged from representation learning \cite{bengio2013representation}. Based on this compact representation, we can measure semantic similarity of words using geometrical properties of the word vector representation, typically the cosine distance between word vectors.

Continuing the work of \cite{florou2018neural}, we are using fastText \cite{joulin2016bag} embeddings trained in the Corpus of Greek Texts \cite{goutsos2010corpus}.

FastText\footnote{fastText, https://fasttext.cc/}, as described in \cite{joulin2016bag} is an efficient library for represantion learning and text classification. Similar to word2vec \cite{mikolov2013distributed}, it produces word embeddings by training a neural language model, that is trying to predict words given context (CBOW architecture) or context given words (SkipGram architecture).
As in word2vec, fastText operates as a neural language model. The key difference with fastText, however, is that it is taking into account morphology, in the form of ngram representations. The representation of a word is calculated as the sum of the embeddings of its ngrams. The ability of fastText to capture morphological information in the produced representations seems to be more efficient compared to other models in downstream tasks  such as text classification.

\section{Data and Methodology}

We trained fastText embeddings on the Corpus of Greek Texts \cite{goutsos2010corpus}, for dimensions ranging from $D=50$ to $D=500$, in steps of 50. The Corpus of Greek Texts consists of approximately 28 million words, a reasonable corpus size to produce meaningful embeddings, able to capture semantic similarity.
FastText is using sub-word information to learn distributed word representation and empirically performs better in downstream NLP tasks compared to word2vec \cite{mikolov2013distributed} or Glo.Ve. \cite{Pennington14glove:global}.
Also, it tackles naturally the problem of spelling errors, as the word-level embeddings are essentially averages of n-gram level embeddings, and words with simple spelling errors still produce very similar embeddings to the intended word.\footnote{fastText embeddings for Greek can be dowloaded here http://sek.edu.gr}

The metaphor  training set consists of 1145 labelled sentences, 563 metaphoric and 582 literal ones. The median length of the words in the training set is 12, minimum number of words is 2, maximum is 225.

To customize the training set we distinguished the phrases between literal and metaphorical according to the Metaphor Identification Procedure
(MIP) as is suggested by the Pragglejaz Group \cite{steen2007finding}. Based on MIP we created two lists of phrases, one literal and another one metaphorical, from the Corpus of Greek Texts. Both of the lists had the same verbs as a kernel but each one could take various objects as predicates. Our training set included some cases of intransitive verbs but did not include collocations, auxiliary, linking, modal or delexical verbs. Furthermore, it must be mentioned that the phrases of our training corpus did not have any metaphor markers which could signal the metaphorical use of a term. Finally, it must be emphasized that in many cases the metaphorical tension is based on the comparison between a human activity and the implementation of the same activity by a non-human.

In  \cite{florou2018neural} the classification is performed by locating the verb in the sentence and averaging the embeddings of a small, fixed-size window centered on the verb, to produce a fixed size input vector for the machine learning algorithm. The averaged context representation is then fed to a Support Vector Machine, which results to 0.83 classification accuracy. The idea of averaging context embeddings in small window sizes comes from \cite{mikolov2013distributed} and the window size is determined empirically.

Here, we extend the fixed-size  contextual representation and, instead, we are passing the entire sentence to the classifier. The classifier then models the probability of a sentence being a metaphor, e.g. $p( label = metaphor | sentence)$ and we optimize the model accordingly.

Both CNNs and RNNs are utilizing the learned (or finetuned) representations of all words in the sentence. This is done by the convolution operator in CNNs and the hidden states in LSTM and GRU reccurent neural networks. Eventually, in both cases, a representation of all words in the sentence is passed to a fully connected layer of the classifier. This improves classification quality, whereas in the simple window-based averaging method, contextual information is distorted for context size larger than 3 or 4 words.

We evaluate all our models with 10-fold cross validation and we report  average accuracy and f1-score.

\subsection{CNN architecture}

The CNN architecture is based on the work of \cite{kim2014convolutional}. More specifically, we are using kernel heights with sizes $k=\{ 3,4, 5\}$ and out channel size of 32. The convolution channels are then max-pooled, concatenated and passed into a fully connected layer. The network is regularized to prevent overfitting by using dropout \cite{srivastava2014dropout}, e.g. dropping units from the network to prevent overfitting, with dropout  probability $0.5$.

\begin{figure}[t]
\includegraphics[width=9cm]{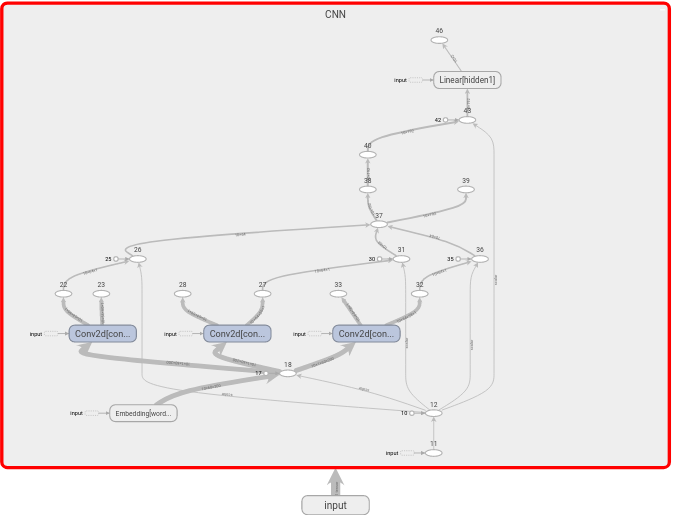}
\centering
\begin{center}
\caption{CNN architecture}
\end{center}
\end{figure}

\subsection{RNN architecture}

In our experiments we tested  Gated Recurrent Units (GRUs, \cite{chung2014empirical}) and Long Short Term Memory architectures (LSTMs, \cite{hochreiter1997long}), using both  unidirectional and bidirectional \cite{schuster1997bidirectional} architectures.

Bidirectional recurrent neural networks are essentially trained on the same sequence of data in forward and backward directions simultaneously and so the output state at every step encodes information about the past (forward direction) and the future (backward direction).

The architecture is exactly the same in both GRU and LSTM configurations, with the recurrence mechanism as the only difference. We are feeding a fixed size, zero padded sentence into the network, followed by the recurrence unit. We then apply 1-max-pooling\footnote{In our experiments we also tried average pooling, with good but inferior results compared to max-pooling, a result  consistent with \cite{huang2015bidirectional} } over the intemmediate hidden layers, followed by a fully connected layer of 100 units and finally the output sigmoid unit.

\begin{figure}[t]
\includegraphics[width=9cm,height=8cm]{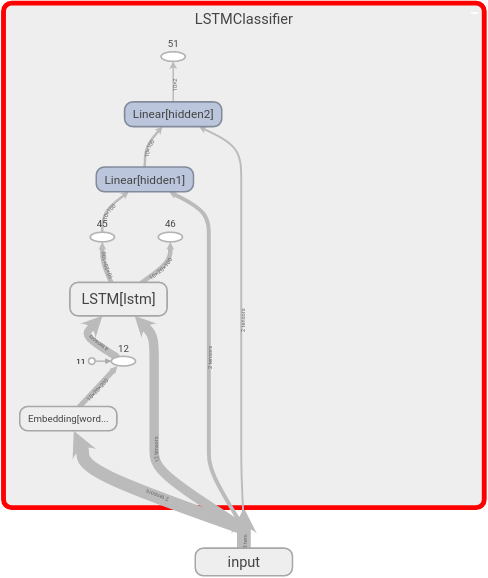}
\centering
\begin{center}
\caption{LSTM architecture}
\end{center}
\end{figure}

\subsection{CRNN}

Finally, we also evaluated a combination of Convolutional Neural Nets and Recurrent Neural Nets, and more specifically the architecture described in \cite{lai2015recurrent}.

Here, the architecture utilises recurrent structure to capture contextual information as far as possible when learning word representations, followed by  a max-pooling layer. Essentially, max-pooling determines which are the most significant words in the underlying text classification problem. Bi-directional architectures consistently outperform uni-directional so we omit results. This is in agreement with \cite{graves2012supervised}.

All  network architectures presented in this paper  are optimised by the Adam optimizer \cite{kingma2014adam}  under the Maximum Likelihood principle and  Negative Log Likelihood as loss function. The implementation is based in PyTorch \cite{paszke2017automatic}. The results of the experiments are summarized in table \ref{table:results}

\begin{figure}[t]
\includegraphics[width=9cm,height=8cm]{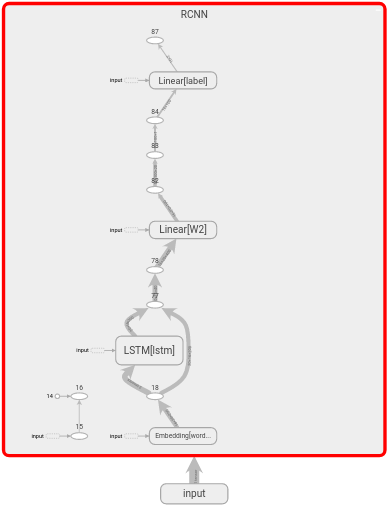}
\centering
\begin{center}
\caption{RCNN architecture}
\end{center}
\end{figure}

\begin{table}[t]
\renewcommand\thetable{1}
\centering

\begin{tabular}{ |p{5cm}||p{1cm}|p{1cm}|  }
 \hline
 \multicolumn{3}{|c|}{Results} \\

 \hline
 Model   & Accuracy  & F1-score \\
 \hline
Florou et.al. 2018 &  0.83  & 0.83 \\
\hline
CNN  ($D=500$) & 0.90 & 0.89 \\
CNN, fine-tuning ($D=150$) & \textbf{0.92} & \textbf{0.92} \\
\hline
b-LSTM  ($D=350)$ & 0.90 & 0.91 \\
b-LSTM, fine-tuning ($D=200)$ & \textbf{0.92} & \textbf{0.92} \\
\hline
b-GRU, fine-tuning ($D=450)$ & 0.91 & 0.91 \\
b-GRU ($D=200)$ & 0.86 & 0.83 \\
\hline
CRNN, fine-tuning ($D=450$)  & 0.91 & 0.91 \\
CRNN, ($D=450)$ & 0.90 & 0.91 \\
\hline

\end{tabular}
\caption{Experiment results, with model architecture and fastText embedding dimentionality}
\label{table:results}

\end{table}

\section{Discussion}

We presented a collection of state-of-the art metaphor detection models achieving accuracy higher than 90\% for the Greek language. This extends the work of \cite{florou2018neural} and, to the best of our knowledge, sets a new state-of-the-art for metaphor detection in Greek, dealing simultaneously with the lack of linguistic resources for Greek. We aim at continuing our work by exploring the performance of contextual embeddings such as ELMO \cite{peters2018deep} and BERT \cite{devlin2018bert}. Another recent promising direction, especially for small datasets is Graph Neural Networks (GNNs) \cite{kipf2017semi,yao2019graph}. In this specific variation of graph neural networks, the entire training set is represented as graph $G=(V,E)$ and the task of the model is node representation and classification, even with potentially few training examples. This is achieved by exploiting the graph structure and the representation of adjacent nodes in the graph.

Both CNNs and bi-directional LSTMs with fine-tuning achieve accuracy higher than 90\%. If we disable fine-tuning, classification accuracy is still high, although  overall fine-tuning  appears to consistently outperform non fine-tuning configurations, which is also consistent with the results presented in \cite{graves2012supervised}.

There are several factors that can explain the performance achieved with neural networks. First, the full sentence is passed into the classifier and thus the model can benefit by exploiting potential long-term semantic dependencies. These dependecies are captured  by the LSTM cells and the convolutional operators. Additionaly, in the case of LSTMs and GRUs, bidirectional architectures appear to consistenly outperform unidirectional architectures.

Finally, transfer learning, in the form of pre-trained embeddings such as fastText  is extremelly useful in the sense that the learned representations capture semantic properties of words in a unsupervised learning fashion and we also allow fine-tuning, which is proven to further enchance the accuracy of the models \cite{Goodfellow-et-al-2016}. Fasttexts' ability to implicitly utilize morphological structure in the form of sub-word representations is also proven to help the overall downstream architecture to significantly improve. We conjecture that this property holds in languages with a rich morphological structure like Greek.

Since it is possible to distinguish between different kinds of metaphor and even between levels of metaphoricity of a term of a sentence, our effort is solely aimed at distinguishing between the literal and the metaphorical use of a term in a specific linguistic context. In that regard, we have not checked at all whether neural language
models have the appropriate properties in order to discriminate pure metaphor from other kinds of
figurative speech such as personification, metonymy, synecdoche etc. In addition, our approach to metaphor detection is not able to classify metaphorical phrases into categories like direct and indirect, or implied and extended. Of course, such an endeavor is a particularly interesting and demanding research challenge, even though the main goal of our specific approach is metaphor detection and its discrimination from literal cases by the use of machine learning algorithms.

\bibliography{references}

\begin{thebibliography}{10}

\bibitem{steen2010metaphor}
Gerard~J Steen, Aletta~G Dorst, J~Berenike Herrmann, Anna~A Kaal, and Tina
  Krennmayr.
\newblock Metaphor in usage.
\newblock {\em Cognitive Linguistics}, 21(4):765--796, 2010.

\bibitem{met_dect_term_relev}
M.~Schulder and E.~D. Hovy.
\newblock Metaphor detection through term relevance.
\newblock In {\em Proceedings of the Second Workshop on Metaphor in NLP.
  Association for Computational Linguistics}, pages 18--26, Baltimore, MD, USA,
  2014.

\bibitem{met_probabilities}
T.~B. Sardinha.
\newblock {\em Metaphor probabilities in corpora. In Zanotto, Mara Sophia,
  Cameron, Lynne and Cavalcanti, Marilda do Couto (eds.) \emph {Confronting
  metaphor in use}}.
\newblock John Benjamins, Amsterdam/Philadelphia, 2008.

\bibitem{Birke}
J.~Birke and A.~Sarkar.
\newblock A clustering approach for the nearly unsupervised recognition of
  nonliteral language.
\newblock In {\em Proc. of the 11th Conference of the European Chapter of the
  Association for Computational Linguistics (EACL-06)}, pages 329--336, Trento,
  Italy, 2006.

\bibitem{metaphor_id}
J.~E. Dunn.
\newblock Evaluating the premises and results of four metaphor identification
  systems.
\newblock In {\em Proceedings of the 14th International Conference on
  Computational Linguistics and Intelligent Text Processing - Volume 2
  (CICLing’13)}, pages 471--486, Samos, Greece, 2013.

\bibitem{dunn2014measuring}
Jonathan Dunn.
\newblock Measuring metaphoricity.
\newblock In {\em Proceedings of the 52nd Annual Meeting of the Association for
  Computational Linguistics (Volume 2: Short Papers)}, pages 745--751, 2014.

\bibitem{blei2003latent}
David~M Blei, Andrew~Y Ng, and Michael~I Jordan.
\newblock Latent dirichlet allocation.
\newblock {\em Journal of machine Learning research}, 3(Jan):993--1022, 2003.

\bibitem{bengio2013representation}
Yoshua Bengio, Aaron Courville, and Pascal Vincent.
\newblock Representation learning: A review and new perspectives.
\newblock {\em IEEE transactions on pattern analysis and machine intelligence},
  35(8):1798--1828, 2013.

\bibitem{koper-etal-2017-ims}
Maximilian K{\"o}per, Evgeny Kim, and Roman Klinger.
\newblock {IMS} at {E}mo{I}nt-2017: Emotion intensity prediction with affective
  norms, automatically extended resources and deep learning.
\newblock In {\em Proceedings of the 8th Workshop on Computational Approaches
  to Subjectivity, Sentiment and Social Media Analysis}, pages 50--57,
  Copenhagen, Denmark, September 2017. Association for Computational
  Linguistics.

\bibitem{rei2017grasping}
Marek Rei, Luana Bulat, Douwe Kiela, and Ekaterina Shutova.
\newblock Grasping the finer point: A supervised similarity network for
  metaphor detection.
\newblock {\em arXiv preprint arXiv:1709.00575}, 2017.

\bibitem{lafferty2001conditional}
John Lafferty, Andrew McCallum, and Fernando~CN Pereira.
\newblock Conditional random fields: Probabilistic models for segmenting and
  labeling sequence data.
\newblock 2001.

\bibitem{kim2014convolutional}
Yoon Kim.
\newblock Convolutional neural networks for sentence classification.
\newblock {\em arXiv preprint arXiv:1408.5882}, 2014.

\bibitem{tang2015document}
Duyu Tang, Bing Qin, and Ting Liu.
\newblock Document modeling with gated recurrent neural network for sentiment
  classification.
\newblock In {\em Proceedings of the 2015 conference on empirical methods in
  natural language processing}, pages 1422--1432, 2015.

\bibitem{wang2016combination}
Xingyou Wang, Weijie Jiang, and Zhiyong Luo.
\newblock Combination of convolutional and recurrent neural network for
  sentiment analysis of short texts.
\newblock In {\em Proceedings of COLING 2016, the 26th international conference
  on computational linguistics: Technical papers}, pages 2428--2437, 2016.

\bibitem{collobert2011natural}
Ronan Collobert, Jason Weston, L{\'e}on Bottou, Michael Karlen, Koray
  Kavukcuoglu, and Pavel Kuksa.
\newblock Natural language processing (almost) from scratch.
\newblock {\em Journal of machine learning research}, 12(Aug):2493--2537, 2011.

\bibitem{florou2018neural}
Eirini Florou, Konstantinos Perifanos, and Dionysis Goutsos.
\newblock Neural embeddings for metaphor detection in a corpus of greek texts.
\newblock In {\em 2018 9th International Conference on Information,
  Intelligence, Systems and Applications (IISA)}, pages 1--4. IEEE, 2018.

\bibitem{joulin2016bag}
Armand Joulin, Edouard Grave, Piotr Bojanowski, and Tomas Mikolov.
\newblock Bag of tricks for efficient text classification.
\newblock {\em arXiv preprint arXiv:1607.01759}, 2016.

\bibitem{goutsos2010corpus}
Dionysis Goutsos.
\newblock The corpus of greek texts: A reference corpus for modern greek.
\newblock {\em Corpora}, 5(1):29--44, 2010.

\bibitem{mikolov2013distributed}
Tomas Mikolov, Ilya Sutskever, Kai Chen, Greg~S Corrado, and Jeff Dean.
\newblock Distributed representations of words and phrases and their
  compositionality.
\newblock In {\em Advances in neural information processing systems}, pages
  3111--3119, 2013.

\bibitem{Pennington14glove:global}
Jeffrey Pennington, Richard Socher, and Christopher~D. Manning.
\newblock Glove: Global vectors for word representation.
\newblock In {\em In EMNLP}, 2014.

\bibitem{steen2007finding}
Gerard~J Steen.
\newblock {\em Finding metaphor in grammar and usage: A methodological analysis
  of theory and research}, volume~10.
\newblock John Benjamins Publishing, 2007.

\bibitem{srivastava2014dropout}
Nitish Srivastava, Geoffrey Hinton, Alex Krizhevsky, Ilya Sutskever, and Ruslan
  Salakhutdinov.
\newblock Dropout: a simple way to prevent neural networks from overfitting.
\newblock {\em The journal of machine learning research}, 15(1):1929--1958,
  2014.

\bibitem{chung2014empirical}
Junyoung Chung, Caglar Gulcehre, KyungHyun Cho, and Yoshua Bengio.
\newblock Empirical evaluation of gated recurrent neural networks on sequence
  modeling.
\newblock {\em arXiv preprint arXiv:1412.3555}, 2014.

\bibitem{hochreiter1997long}
Sepp Hochreiter and J{\"u}rgen Schmidhuber.
\newblock Long short-term memory.
\newblock {\em Neural computation}, 9(8):1735--1780, 1997.

\bibitem{schuster1997bidirectional}
Mike Schuster and Kuldip~K Paliwal.
\newblock Bidirectional recurrent neural networks.
\newblock {\em IEEE Transactions on Signal Processing}, 45(11):2673--2681,
  1997.

\bibitem{huang2015bidirectional}
Zhiheng Huang, Wei Xu, and Kai Yu.
\newblock Bidirectional lstm-crf models for sequence tagging.
\newblock {\em arXiv preprint arXiv:1508.01991}, 2015.

\bibitem{lai2015recurrent}
Siwei Lai, Liheng Xu, Kang Liu, and Jun Zhao.
\newblock Recurrent convolutional neural networks for text classification.
\newblock In {\em Twenty-ninth AAAI conference on artificial intelligence},
  2015.

\bibitem{graves2012supervised}
Alex Graves.
\newblock Supervised sequence labelling with recurrent neural networks. 2012.
\newblock {\em URL http://books. google. com/books}, 2012.

\bibitem{kingma2014adam}
Diederik~P Kingma and Jimmy Ba.
\newblock Adam: A method for stochastic optimization.
\newblock {\em arXiv preprint arXiv:1412.6980}, 2014.

\bibitem{paszke2017automatic}
Adam Paszke, Sam Gross, Soumith Chintala, Gregory Chanan, Edward Yang, Zachary
  DeVito, Zeming Lin, Alban Desmaison, Luca Antiga, and Adam Lerer.
\newblock Automatic differentiation in pytorch.
\newblock 2017.

\bibitem{peters2018deep}
Matthew~E Peters, Mark Neumann, Mohit Iyyer, Matt Gardner, Christopher Clark,
  Kenton Lee, and Luke Zettlemoyer.
\newblock Deep contextualized word representations.
\newblock {\em arXiv preprint arXiv:1802.05365}, 2018.

\bibitem{devlin2018bert}
Jacob Devlin, Ming-Wei Chang, Kenton Lee, and Kristina Toutanova.
\newblock Bert: Pre-training of deep bidirectional transformers for language
  understanding.
\newblock {\em arXiv preprint arXiv:1810.04805}, 2018.

\bibitem{kipf2017semi}
Thomas~N. Kipf and Max Welling.
\newblock Semi-supervised classification with graph convolutional networks.
\newblock In {\em International Conference on Learning Representations (ICLR)},
  2017.

\bibitem{yao2019graph}
Liang Yao, Chengsheng Mao, and Yuan Luo.
\newblock Graph convolutional networks for text classification.
\newblock In {\em Proceedings of the AAAI Conference on Artificial
  Intelligence}, volume~33, pages 7370--7377, 2019.

\bibitem{Goodfellow-et-al-2016}
Ian Goodfellow, Yoshua Bengio, and Aaron Courville.
\newblock {\em Deep Learning}.
\newblock MIT Press, 2016.
\newblock http://www.deeplearningbook.org.

\end{thebibliography}

\end{document}